\documentclass[letterpaper]{article} 
\usepackage[]{aaai24}  
\usepackage{times}  
\usepackage{helvet}  
\usepackage{courier}  
\usepackage[hyphens]{url}  
\usepackage{graphicx} 
\usepackage{natbib}  
\usepackage{caption} 
\usepackage{algorithm}
\usepackage{algorithmic}
\usepackage{amsmath}
\usepackage{amssymb}
\usepackage{amsthm}
\usepackage{xcolor}
\usepackage{booktabs}
\usepackage{multirow}
\usepackage{array} 
\usepackage[utf8]{inputenc}
\usepackage{color}
\usepackage{enumitem}
\newlist{checklist}{itemize}{1}
\setlist[checklist,1]{
  label={\(\square\)},
  labelwidth=1em,
  labelsep=1em,
  left=0pt,
  itemsep=0pt
}
\usepackage{newfloat}
\usepackage{listings}
\DeclareCaptionStyle{ruled}{labelfont=normalfont,labelsep=colon,strut=off} 
\floatstyle{ruled}
\newfloat{listing}{tb}{lst}{}
\floatname{listing}{Listing}
\usepackage{hyperref}
\nocopyright

\title{Orthogonal Finetuning for Direct Preference Optimization}
\author{
    Chenxu Yang\textsuperscript{\rm 1,2}\thanks{\ \ \ Equal contribution. }, Ruipeng Jia\textsuperscript{\rm 3}\footnotemark[1],
    Naibin Gu\textsuperscript{\rm 1,2}, Zheng Lin\textsuperscript{\rm 1,2}\thanks{\ \ \ Zheng Lin is the corresponding author. }, Siyuan Chen\textsuperscript{\rm 1,2},\\
{\bf Chao Pang\textsuperscript{\rm 3}, Weichong Yin\textsuperscript{\rm 3}, Yu Sun\textsuperscript{\rm 3}, Hua Wu\textsuperscript{\rm 3}, Weiping Wang\textsuperscript{\rm 1}
}}
\affiliations{
   \textsuperscript{\rm 1}Institute of Information Engineering, Chinese Academy of Sciences, Beijing, China \\
  \textsuperscript{\rm 2}School of Cyber Security, University of Chinese Academy of Sciences, Beijing, China \\
  \textsuperscript{\rm 3}Baidu Inc., Beijing, China\\
  \textrm{\{}yangchenxu,linzheng,wangweiping\textrm{\}}@iie.ac.cn \\
  {jiaruipeng@baidu.com}
}

\usepackage{bibentry}

\begin{document}

\maketitle

\begin{abstract}
DPO is an effective preference optimization algorithm. However, the DPO-tuned models tend to overfit on the dispreferred samples, manifested as overly long generations lacking diversity. While recent regularization approaches have endeavored to alleviate this issue by modifying the objective function, they achieved that at the cost of alignment performance degradation. In this paper, we innovatively incorporate regularization from the perspective of weight updating to curb alignment overfitting. Through the pilot experiment, we discovered that there exists a positive correlation between overfitting and the hyperspherical energy fluctuation. Hence, we introduce orthogonal finetuning for DPO via a weight-\textbf{Ro}tated \textbf{P}reference \textbf{O}ptimization (\textbf{RoPO}) method, which merely conducts rotational and magnitude-stretching updates on the weight parameters to maintain the hyperspherical energy invariant, thereby preserving the knowledge encoded in the angle between neurons. Extensive experiments demonstrate that our model aligns perfectly with human preferences while retaining the original expressive capacity using only 0.0086\% of the trainable parameters, suggesting an effective regularization against overfitting. Specifically, RoPO outperforms DPO by up to 10 points on MT-Bench and by up to 2.8 points on AlpacaEval 2, while enhancing the generation diversity by an average of 6 points.
\footnote{ The code is released at https://github.com/iie-ycx/ropo.}
\end{abstract}

\section{Introduction}

While large language models (LLM) have achieved astonishing performance \cite{chatgpt,touvron2023llama2,bai2023qwen,yang2023baichuan}, they still encounter risks of generating content undesirable from the human perspective \cite{bai2022constitutionalaiharmlessnessai}. 
Consequently, reinforcement learning from human's feedback (RLHF) was introduced to ensure controllable AI systems by mimicking human preferences among multiple candidate answers \cite{christiano2023deepreinforcementlearninghuman,ouyang2022traininglanguagemodelsfollow,stiennon2022learningsummarizehumanfeedback}. However, RLHF are notorious for its training instability and sensitivity to hyperparameters. Recently, some researchers designed RL-free direct alignment algorithms \cite{dong2023raftrewardrankedfinetuning,yuan2023rrhfrankresponsesalign,zhao2023slichfsequencelikelihoodcalibration}, and Direct Preference Optimization (DPO) is a representative work in this domain \cite{rafailov2023directpreferenceoptimizationlanguage}. Derived with the aim of attaining an optimum of the KL-constrained reward maximization, DPO circumvents the explicit modeling of the reward model and unstable reinforcement learning through reparameterization, optimizing the policy by employing the cross entropy objective on pairwise data. The reverse KL divergence regularization in the objective is designed to ensure that new desirable behaviors are learned without losing the expressiveness and fluency of the original model, avoiding the problem of reward hacking \cite{azar2023generaltheoreticalparadigmunderstand,zeng2024tokenleveldirectpreferenceoptimization}.

\begin{figure}[!t]
  \centerline{\includegraphics[scale=0.4]{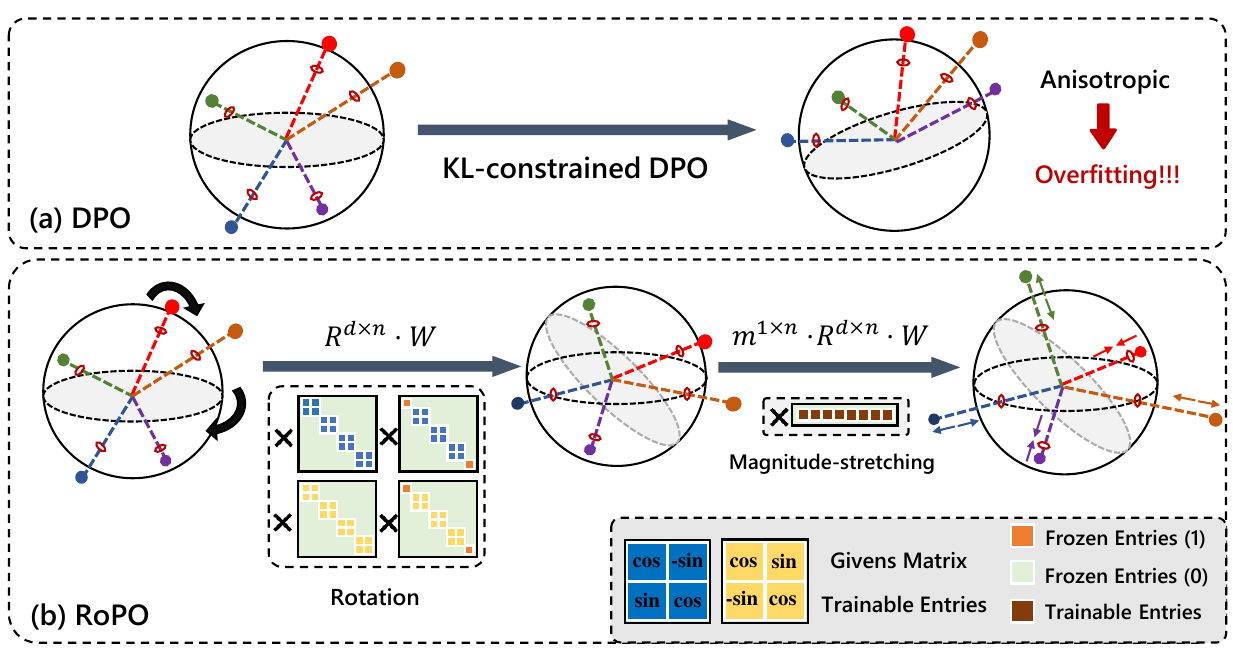}}
  \caption{Comparison of original DPO and our method RoPO. The original DPO training led to the change of the angle between neurons. The isotropic distribution of the original neurons in some layers of the model was disrupted, thereby causing overfitting. The rotational and stretching weight update approach of RoPO in the alignment training effectively avoids this issue and preserves the knowledge encoded in the relative positions of neurons.}
  \label{figure1}
\end{figure} 

Unfortunately, a fatal defect exists in DPO:
\textbf{it causes the model to overfit on the behavior of suppressing dispreferred examples}, as the model has to push the probability of dispreferred sample as close to 0 as possible to maximize the DPO objective. The overfitting issue wrongly restricts some useful characteristics in the dispreferred examples (eg. the generation length expands abnormally due to the mistaken suppression of the termination token $<$eos$>$), eventually leading to the generation of the DPO-tuned model being overly lengthy and lacking diversity. Numerous approaches have been proposed to alleviate this problem through modifying the objective function \cite{azar2023generaltheoreticalparadigmunderstand,wang2023reverseklgeneralizingdirect,zeng2024tokenleveldirectpreferenceoptimization}; however, they achieved that at the cost of alignment performance degradation.

Given that the motivation for introducing reverse KL divergence into DPO is to prevent the policy model from deviating too far from the reference model, we associate that we could attempt to incorporate regularization from the perspective of weight updating to achieve the same effect. Specifically, we hope to design an effective weight update strategy to mitigate the influence of the gradient update on the model parameters when overfitting emerges and preserve the knowledge acquired in the previous training stage. 
\citet{qiu2024controllingtexttoimagediffusionorthogonal} proposed that the angle between neurons represents the knowledge of the neural network, and maintaining the uniform distribution of neurons during fine-tuning can maximize the semantic generation capacity of the model \cite{liu2020learningminimumhypersphericalenergy}. Inspired by them, we hypothesized that the overfitting issue of DPO could be attributed to the reduction in the uniformity of the arrangement of neurons on the unit hypersphere as depicted in Figure \ref{figure1}, where neurons tend to cluster in a dense space together. To validate this, we designed pilot experiments to observe the changes of the hyperspherical energy value, which indicates the diversity of neurons, before and after DPO training. The experimental results exhibit an increase in the hyperspherical energy in the partial mid and high layers of neuron networks, suggesting that the original isotropic arrangement property of neurons was disrupted. Therefore, we proposed the weight-\textbf{Ro}tated \textbf{P}reference \textbf{O}ptimization (\textbf{RoPO}) method, which merely conducts rotational and magnitude-stretching updates on the weight parameters of the policy model to retain the relative angles between paired neurons. Under such weight-updating constraints, {RoPO} preserves the knowledge encoded in the relative positions of neurons.
Extensive experiments reveal that {RoPO} achieves a performance on the alignment task comparable to that of the strongest baseline with merely 0.0086\% of the trainable parameters, while effectively suppressing the overfitting phenomenon, as manifested by the preservation of diverse expressions, normal generation length, and no obvious knowledge forgetting.

Our contributions are summarized as follows:
 \begin{itemize}
	\item We conducted a systematic analysis of the causes of overfitting induced by DPO from multiple perspectives.
	\item We proposed the design of regularization from the parameter perspective to alleviate the overfitting problem of DPO. To the best of our knowledge, our RoPO is the first to adopt this approach.
	\item RoPO has performed outstandingly on multiple evaluation benchmarks. It has achieved a good balance between the alignment performance and expression ability. Additionally, RoPO has significantly reduced the number of training parameters and enhanced training efficiency.
\end{itemize}

\section{Preliminaries}

\subsection{Direct Preference Optimization}
DPO is an optimization method that directly learns the policy bypassing the reward function. \citet{rafailov2023directpreferenceoptimizationlanguage} derived the optimal solution of the reward function r* based on the original RL objective function as:
\begin{equation}
r^*(x,y)=\beta\log\frac{\pi_\theta(y\mid x)}{\pi_{\mathrm{ref}}(y\mid x)}+\beta\log Z(x),
\end{equation}
where $\pi_\theta$ is the policy model, $\pi_{\mathrm{ref}}$ is the policy model, and $Z(x)$ denotes the partition function.

\begin{equation}
p^*(y_1\succ y_2|x)=\frac{\exp\left(r^*(x,y_1)\right)}{\exp\left(r^*(x,y_1)\right)+\exp\left(r^*(x,y_2)\right)}.
\end{equation}
Subsequently, they incorporated it into the Bradley-Terry (BT) \cite{BTmodelBradley1952RankAO} model, eliminated the partition function, defined the objective function as the maximum likelihood of $p^*$, and ultimately obtained:
\begin{equation}
\begin{aligned}
&\mathcal{L}_{\mathrm{DPO}}(\pi_\theta;\pi_{\mathrm{ref}})=\\
&-\mathbb{E}\left[\log\sigma\left(\beta\log\frac{\pi_\theta(y_w\mid x)}{\pi_{\mathrm{ref}}(y_w\mid x)}-\beta\log\frac{\pi_\theta(y_l\mid x)}{\pi_{\mathrm{ref}}(y_l\mid x)}\right)\right],
\end{aligned}
\end{equation}
where $x$ denotes the prompt, $y_w$ denotes the winning response, $y_l$ denotes the losing response.

\subsection{Hyperspherical Energy}
The initial proposal of Hyperspherical Energy (HE) was motivated by the diversification and balanced distribution of neurons to prevent the parameter redundancy problem \cite{liu2020learningminimumhypersphericalenergy}. Inspired by the renowned physics problem known as Thomson problem, \citet{liu2020learningminimumhypersphericalenergy} designed the neural network training objective with Minimum Hyperspherical Energy (MHE) as the regularization. Assuming that there is a fully connected layer $\boldsymbol{W=}\{\boldsymbol{w}_1,\cdots,\boldsymbol{w}_n\}\boldsymbol{\in}\mathbb{R}^{d\times n}$, where $\boldsymbol{w}_i\in\mathbb{R}^d$ denotes the $i$-th neuron. The definition of HE is as follows:

\begin{equation}
{\mathrm{HE}}(\boldsymbol{W}) = \sum_{i\neq j}\|\hat{\boldsymbol{w}}_i-\hat{\boldsymbol{w}}_j\|^{-1}
\end{equation}
where $\boldsymbol{\hat{w}}_i=\boldsymbol{w}_i/\|\boldsymbol{w}_i\|$ is the $i$-th normalized neuron.

\subsection{Givens Rotation}

The Givens matrix is a commonly used rotation matrix. It rotates a vector in a 2-dimensional subspace planes, and the rotation angle is controlled by $\theta$. Supposing we have the following Givens matrix, where $\cos\theta$ appears at $\{(i, i), (j, j)\}$, $\sin\theta$ appears at $\{(i, j), (j, i)\}$. In this paper, we regarded the Givens rotation matrix as the minimum unit for constructing orthogonal matrices of RoPO.
\begin{equation}
G(i,j,\theta)=\begin{pmatrix}I&0&0&0&0\\0&\cos\theta&0&\sin\theta&0\\0&0&I&0&0\\0&-\sin\theta&0&\cos\theta&0\\0&0&0&0&I\end{pmatrix}
\end{equation}
where $I$ denotes identity matrix.

\begin{figure}[!t]
  \centerline{\includegraphics[scale=0.27]{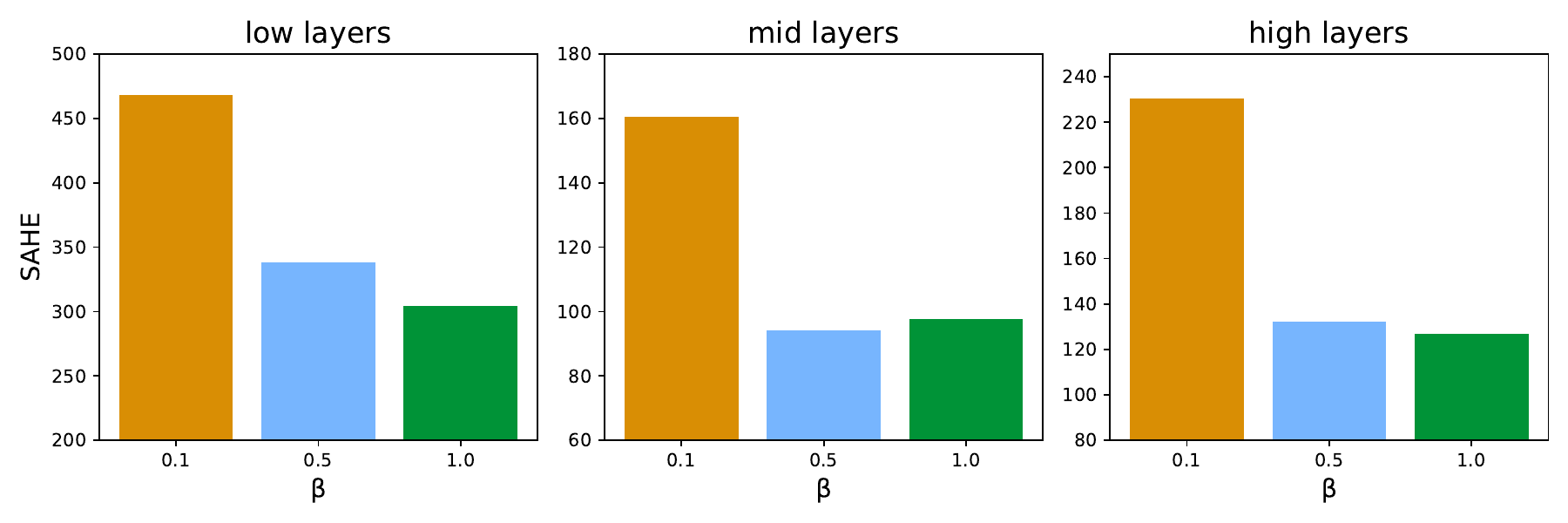}}
  \caption{The summation of absolute hyperspherical energy variation (SAHE) across layers in LLM after DPO training.}
  \label{figure2}
\end{figure} 

\begin{figure}[!t]
  \centerline{\includegraphics[scale=0.27]{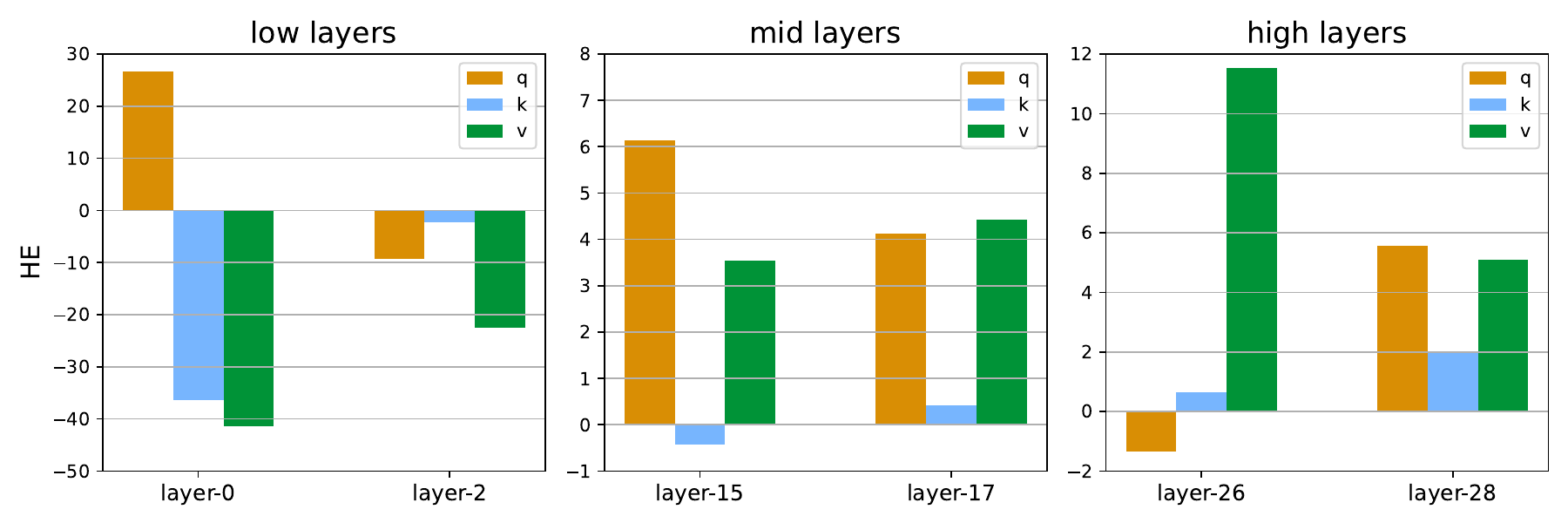}}
  \caption{The hyperspherical energy (HE) variations of different networks across different layers in LLM after DPO training ($\beta=0.1$). $q$ denotes query vector, $k$ denotes key vector, and $v$ denotes value vector in attention mechanism.}
  \label{figure3}
\end{figure}

\section{Alignment Overfitting in DPO}
\subsection{Theoretical analysis}
In accordance with the Bradley-Terry (BT) preference model, \citet{rafailov2023directpreferenceoptimizationlanguage} formulated the probability that the positive example surpasses the dispreferred example in each sample pair as:
\begin{equation}
p^*(y_w\succ y_l|x) =\sigma\left(\beta\log\frac{\pi_\theta(y_w|x)}{\pi_{\mathrm{ref}}(y_w|x)}-\beta\log\frac{\pi_\theta(y_l|x)}{\pi_{\mathrm{ref}}(y_l|x)}\right).
\end{equation}

The probability could be re-written as follows:
\begin{equation}
p^*(y_w\succ y_l|x)= \sigma\left(\beta\log\frac{\pi_\theta(y_w\mid x)}{\pi_\theta(y_l\mid x)}-\gamma\right),
\end{equation}
where $\gamma=\beta\log\frac{\pi_{\mathrm{ref}}(y_w\mid x)}{\pi_{\mathrm{ref}}(y_l\mid x)}$ has no relation to the parameter update during training and can be regarded as a constant.

To minimize the DPO loss function, the model would try to increase the probability $p^*(y_w\succ y_l|x)$, which could be achieved by increasing the ratio $\frac{\pi_\theta(y_w\mid x)}{\pi_\theta(y_l\mid x)}$. Nevertheless, the increase of $\pi_\theta(y_w|x)$ has an upper bound $1$. Hence, the model turns to push the probability of $y_l$ as close to $0$ as possible, which leads the model to overfitting on the behavior of suppressing dispreferred examples. The original intention of alignment was merely to suppress the undesirable behaviors in the dispreferred examples. However, overfitting suppresses all behavioral characteristics, no matter good or bad, resulting in poor expressive ability and the loss of generation diversity. A common issue of DPO is that the generation length expands abnormally throughout the training procedure, which can be attributed to overfitting mistakenly suppressing the termination symbol $<$eos$>$ \cite{dubey2024llama3herdmodels}.

In the DPO objective, $\beta$ governs the deviation from the reference model $\pi_{\mathrm{ref}}$. We believe that through adjusting $\beta$, the alignment intensity can be regulated, thereby controlling overfitting. The following is our explanation. Assuming that the true preference of a sample $p^*(y_w\succ y_l|x)=\hat{p}$, if the $\beta$ value shrinks to  approach $0$, the model has to learn to further reduce the probability of dispreferred completions $\pi_\theta(y_l| x)$ to fit the true preference probability $\hat{p}$; if the $\beta$ value increases, the model merely needs to learn to decrease the probability of dispreferred completions with a relatively weaker strength.

\subsection{Pilot Experiment on Hyperspherical Energy}

Recent studies revealed that maintaining hyperspherical energy unchanged is crucial for preserving the semantic generation capacity of text-to-image diffusion models \cite{qiu2024controllingtexttoimagediffusionorthogonal}. They held the view that some of the model's knowledge is contained within the relative angles between neurons. Inspired by them, we hypothesize that the expressive capacity degradation caused by alignment overfitting problem might also be related to the damage of angular encoded knowledge in neurons during DPO training. To validate it, we devised experiments to observe the hyperspherical energy variations, before and after DPO training.

We initially devised experiments to investigate the relationship between the absolute values of the alterations in hyperspherical energy and the parameter $\beta$ in DPO. The summation of absolute hyperspherical energy variation (SAHE) is utilized to exhibit it, which is calculated as follows:
\begin{equation}
\mathrm{SAHE}(\boldsymbol{W_0},\boldsymbol{W_1},L)=\sum_{l\in L}|
\mathrm{HE}(\boldsymbol{W^l_1})-\mathrm{HE}(\boldsymbol{W^l_0})|.
\end{equation}

The experiments depicted in Figure \ref{figure2} indicate the variation of the hyperspherical energy when the overfitting controlling parameter $\beta$ varies. The experimental results demonstrate a generally positive correlation between them, as evidenced by the fact that intense fluctuations of hyperspherical energy are accompanied by a lower $\beta$ value across all layers.
Then, we observed how the hyperspherical energy of different networks across different layers variates in detail, and corresponding results are shown in Figure \ref{figure3}. The findings highlight that DPO training leads to an increase in the hyperspherical energy of the model in the middle and high layers, and this phenomenon is more pronounced on the query and value vectors. In the low layers, the hyperspherical energy of the model is more unstable, exhibiting significant fluctuation. Hence, we hypothesize that DPO makes the distribution of neurons in the middle and high layers compact, leading to an impaired isotropy; while in the low layers, the disruption of the relative angle is severe, resulting in the loss of corresponding knowledge. 

\begin{figure*}[htbp]
  \centerline{\includegraphics[scale=0.65]{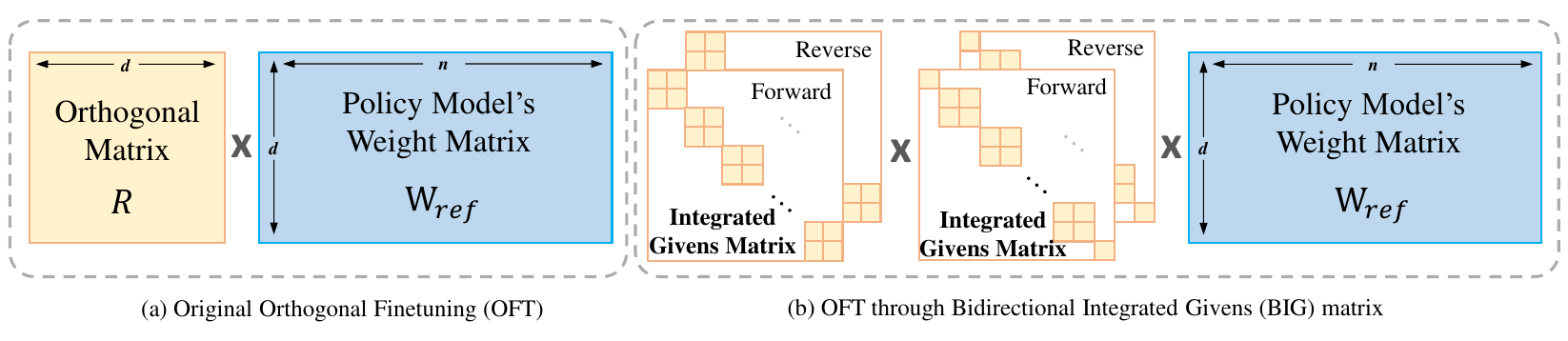}}
  \caption{Comparison of original orthogonal matrix and BIG orthogonal matrix.}
  \label{figure4}
\end{figure*} 
\section{Methodology}

RoPO attempts to incorporate regularization from the perspective of weight update to curb overfitting. This constraint approach is similar to Parameter-Efficient Fine-Tuning (PEFT), and one popular PEFT method is Low-Rank Adaptation (LoRA) \cite{hu2021loralowrankadaptationlarge}:
\begin{equation}
h=W_0x+\Delta Wx=W_0x+BAx,
\end{equation}
where $B\in\mathbb{R}^{d_1\times r}$ and $A\in\mathbb{R}^{r\times d_2}$ with the rank $r \ll min(d_1,d_2)$ are trainable matrices.

Motivated by the investigation in pilot experiments, we devised a weight-regularized training approach that maintains the hyperspherical energy (HE) invariant, named weight-Rotated Preference Optimization (RoPO). RoPO retains the reverse KL divergence constraint design in original DPO and incorporates additional weight-updating constraints to prevent the suppression of useful features in dispreferred completions caused by overfitting. Additionally, it shares the same benefit as LoRA of not introducing additional overhead during the inference stage since the parameter matrices can be merged. The HE-invariant weight regularization conform to the following equation:
\begin{equation}
\left\|\sum_{i\neq j}\|\hat{\boldsymbol{w}}_i-\hat{\boldsymbol{w}}_j\|^{-1}-\sum_{i\neq j}\|\hat{\boldsymbol{w}}_i^0-\hat{\boldsymbol{w}}_j^0\|^{-1}\right\|=0,
\end{equation}
where $\hat{\boldsymbol{w}}_i^0$ denotes the weight before fine-tuning, and $\hat{\boldsymbol{w}}_i$ denotes the weight after fine-tuning.

Performing orthogonal transformation on the weight matrix constitutes a feasible approach for keeping hyperspherical energy invariant. There exist diverse approaches to construct orthogonal matrices, encompassing: Gram-Schmidt method \cite{liu2021orthogonaloverparameterizedtraining}, Cayley parameterization \cite{qiu2024controllingtexttoimagediffusionorthogonal}, Householder Reflection \cite{yuan2024bridginggaplowrankorthogonal}, and Givens Rotation \cite{GOFTma2024parameterefficientquasiorthogonalfinetuning}. They could be summarily expressed as the following equation:
\begin{equation}
{h}={W}{x}=({R}\cdot{W}^0){x},\quad\mathrm{s.t.}{R}^\top{R}={R}{R}^\top={I}
\end{equation}
where $R$ denotes the tunable orthogonal weight matrix and $I$ denotes an identity matrix.

Different from the aforementioned approaches, RoPO adopts Bidirectional Integrated Givens (BIG) matrices $\tilde{G}$ to construct orthogonal matrix $R$.
BIG matrices $\tilde{G}$ consist of two fundamental matrix units, namely the forward Givens rotation matrix $G$ and the reverse Givens rotation matrix $G'$. $G$ could achieve counter-clockwise rotation and $G'$ is capable of attaining clockwise rotation in a 2-dimensional subspace planes. Their construction is presented as follows: 
\begin{equation}
{G}=\begin{bmatrix}\cos\theta&-\sin\theta\\\sin\theta&\cos\theta\end{bmatrix},\quad {G}'=\begin{bmatrix}\cos\theta&\sin\theta\\-\sin\theta&\cos\theta\end{bmatrix}.
\end{equation}

\citet{GOFTma2024parameterefficientquasiorthogonalfinetuning} proved that in a $d$-dimensional linear space, it requires at most $d - 1$ specific Givens rotations to rotate a vector $\boldsymbol{x}\in\mathbb{R}^d $ to any vector $\boldsymbol{y}\in\mathbb{R}^d$ on the same sphere with $\boldsymbol{x}$.  Based on their theory, we organize the matrix units in the manner presented in Figure \ref{figure4} to design the BIG matrices. The specific arrangement of the Givens matrices in BIG guarantees efficient computations, as it only requires four sparse matrix multiplications to obtain a rotation matrix with full-angle coverage\footnote{Although both employ the given rotation to construct orthogonal matrices, RoPO and GOFT \cite{GOFTma2024parameterefficientquasiorthogonalfinetuning} differ in terms of methods and motivations. Not only can RoPO express any rotation with fewer sparse matrix multiplications, but it also introduces an additional trainable vector that governs the length scaling.}.
\begin{equation}
\begin{aligned}
\tilde{G_1}/\tilde{G'_1}&=\prod_{k=0}^{(d/2)-1}{G}/{G}'(2k,2k+1;\theta_k), \\
\tilde{G_2}/\tilde{G'_2}&=\prod_{k=0}^{(d-1)/2}{G}/{G}'(2k+1,2k+2;\theta_k),
\end{aligned}
\end{equation}

Furthermore, in order to explore sufficient optimization spaces for alignment learning, we added a trainable vector $m$ that controls the length scaling in RoPO. It can be observed from Figure \ref{figure1} that the scaling of the neurons' magnitudes still ensures the invariance of the hyperspherical energy. The final weight regularization constraint of RoPO can be expressed by the following formula:
\begin{equation}
{h}=\underline{m}\cdot(\underline{R}\cdot{W^0}){x}=\left(\left(\underline{\tilde{G_1}}\cdot\underline{\tilde{G_2}}\cdot\underline{\tilde{G'_1}}\cdot\underline{\tilde{G'_2}}\right){W^0}\right){x},
\end{equation}
where $W^0$ remains frozen during DPO-tuning, and the underlined parameters are trainable.

Our method could incorporate the rapid implementation approach of matrix multiplication in Rotary Position Embedding (RoPE) \cite{su2023roformerenhancedtransformerrotary}, which prominently accelerates the training process. The implementation of the enhancement scheme is located in the Appendix D.

\begin{table*}[]
\centering
\renewcommand\arraystretch{1.0}
\scalebox{1}{
\begin{tabular}{lcccccccccccc}
\toprule
\multirow{3}{*}{\textbf{Method}} & \multicolumn{6}{c}{\textbf{Mistral-Base (7B)}}                                                                              & \multicolumn{6}{c}{\textbf{Llama3-Base (8B)}}                                                                               \\ \cline{2-13} 
                                 & \multicolumn{2}{c}{\textbf{AlpacaEval 2}} & \multicolumn{2}{c}{\textbf{Arena-Hard}} & \multicolumn{2}{c}{\textbf{MT-Bench}} & \multicolumn{2}{c}{\textbf{AlpacaEval 2}} & \multicolumn{2}{c}{\textbf{Arena-Hard}} & \multicolumn{2}{c}{\textbf{MT-Bench}} \\ \cline{2-13} 
                                 & \textbf{WWR}       & \textbf{Len.}      & \textbf{WWR}      & \textbf{Len.}     & \textbf{WWR}     & \textbf{Len.}    & \textbf{WWR}       & \textbf{Len.}      & \textbf{WWR}      & \textbf{Len.}     & \textbf{WWR}     & \textbf{Len.}    \\ \hline
SFT                              & 4.15                 & 790                & 5.40                & 1157              & 12.81              & 808              & 7.54                 & 842                & 8.70                & 1221              & 14.10              & 834              \\
DPO($\beta$=0.1)                 & 10.19                & 1347               & 14.71               & 1614              & 14.78              & 1591             & 10.75                & 1144               & 15.35               & 1567              & 20.86              & 1060             \\
DPO($\beta$=0.3)                 & 9.27                 & 1194               & 10.85               & 1470              & 13.79              & 1390             & 7.95                 & 1005               & 12.62               & 1436              & 17.23              & 942              \\
IPO                              & 8.75                 & 1007               & 12.99               & 1347              & 11.14              & 929              & 8.36                 & 1054               & 14.26               & 1512              & 15.82              & 1026             \\
KTO                              & 8.92                 & 1286               & 12.92               & 1475              & 14.67              & 1106             & 6.82                 & 1292               & 18.01               & 1665              & 10.30              & 1142             \\
ORPO                             & 5.72                 & 1074               & 9.76                & 1345              & 13.40              & 992              & 5.00                 & 1106               & 12.95               & 1447              & 16.49              & 984              \\
R-DPO                            & 9.93                 & 1273               & 12.56               & 1642              & 9.92               & 1340             & 11.08                & 1128               & 16.06               & 1517              & 13.17              & 1009             \\
LoPO(r=4)                    & 10.01                & 962                & 11.21               & 1421              & 15.98              & 920              & 10.53                & 1012               & 14.34               & 1503              & 12.14              & 969              \\
LoPO(r=16)                   & 10.01                & 983                & 11.63               & 1450              & 17.71              & 996              & 10.04                & 1000               & 16.50               & 1515              & 13.40              & 992              \\
DoPO(r=4)                    & 10.59                & 967                & 14.88               & 1428              & 17.03              & 953              & 9.03                 & 998                & 15.49               & 1513              & 20.70              & 994              \\
RoPO(our)                        & \textbf{13.03}                & 1038               & \textbf{15.85}               & 1419              & \textbf{22.84}              & 968              & \textbf{11.38}                & 1062               & \textbf{18.63}               & 1509              & \textbf{26.20}              & 1010             \\  \bottomrule
\end{tabular}
}
\caption{Evaluation results on three instruction following benchmarks: AlpacaEval 2, Arena-Hard, and MT-Bench. \textbf{WWR} denotes length weighted win rate (\%) and \textbf{Len.} is the abbreviation of avarege generation length. The best results are highlighted with \textbf{bold}.}
\label{table1}
\end{table*}

\section{Experiments}

\subsection{Experimental Setup}

\textbf{Tasks.}
We evaluated our method in two tasks, namely: the open-ended instruction-following task and the commonsense reasoning question-answering tasks. The open-ended instruction-following task encompasses instruction finetuning and preference optimization datasets, and we regarded it as the main experiment for assessing the alignment performance of the model. Additionally, we incorporated the commonsense reasoning tasks from the Huggingface Open Leaderboard benchmarks \cite{open-llm-leaderboard-v1} to evaluate the forgetting problem caused by the overfitting issue in the preference optimization methods.

\noindent \textbf{Training details.} We chose the popular Meta-Llama-3-8B \cite{dubey2024llama3herdmodels} and Mistral-7B-v0.1 model \cite{jiang2023mistral7b} as the backbone model and conducted experiments on them. During the training process, for the open-ended instruction-following task, we instruction-finetune (IFT) a base model on the UltraChat-200k dataset \cite{ding-etal-2023-enhancing} to acquire a SFT model. Subsequently, we conduct preference optimization on the UltraFeedback dataset  \cite{cui2024ultrafeedbackboostinglanguagemodels} using the SFT model as the reference model. The commonsense reasoning question-answer tasks consist of 8 sub-tasks, each of which is equipped with a predefined training and testing set. We merged these 8 training sets into an integrated downstream task training set. In the experiment aimed at evaluating alignment overfitting, we first trained the pretrained base model on this commonsense reasoning QA training set until convergence, and then conducted SFT and preference optimization on the instruction-following dataset. The overfitting problem could be manifested by comparing the influence of different preference optimization methods on the performance of commonsense reasoning tasks as we hypothesized that overfitting would lead to knowledge forgetting. For more training details such as hyperparameter settings, please refer to Appendix A.

\noindent \textbf{Baselines.}
We compare RoPO with the following offline preference optimization methods: DPO \cite{rafailov2023directpreferenceoptimizationlanguage}, IPO \cite{azar2023generaltheoreticalparadigmunderstand}, KTO \cite{KTOethayarajh2024ktomodelalignmentprospect}, ORPO \cite{ORPOhong2024orpomonolithicpreferenceoptimization}, R-DPO \cite{RDPOpark2024disentanglinglengthqualitydirect}. Additionally, we add additional baselines LoRA-regularized DPO (LoPO) and DoRA-regularized DPO (DoPO) to test other feasible weight regularization methods for mitigating the overfitting. The detailed introductions of these baselines are given in the Appendix B.

\begin{figure*}[htbp]
  \centerline{\includegraphics[scale=0.45]{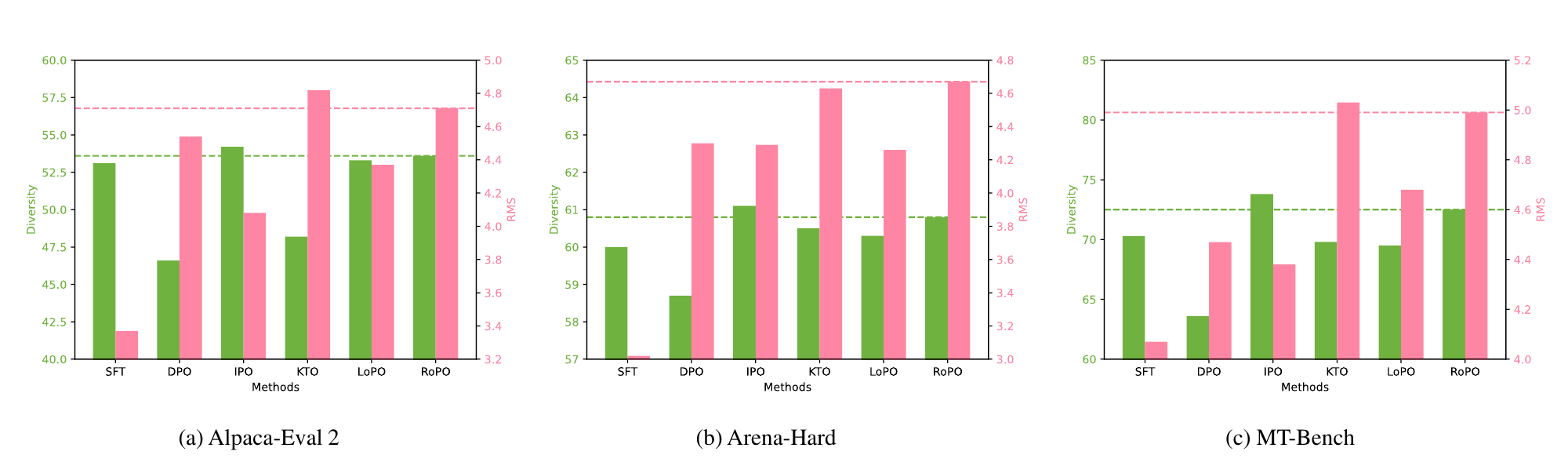}}
  \caption{Reward model score (RMS) and diversity results of the \textbf{Mistral-7B-v0.1} model on the three benchmarks. }
  \label{figure5}
\end{figure*} 
\begin{table*}[]
\centering
\renewcommand\arraystretch{1.0}
\scalebox{1.0}{
\begin{tabular}{lccccccccc}
\toprule
\textbf{Method} & \textbf{BoolQ} & \textbf{PIQA} & \textbf{SIQA} & \textbf{HellaSwag} & \textbf{WinoGrande} & \textbf{ARC-e} & \textbf{ARC-c} & \textbf{OBQA} & \textbf{Avg.} \\ \hline

TS-FT           & 74.22          & 86.72         & 80.45         & 93.98              & 85.95               & 90.15          & {80.46}          & \textbf{88.40}         & {85.04}              \\ \hline
SFT             & 55.11          & 75.24         & 55.53         & 44.85              & 50.99               & 66.37          & 61.09          & 54.00         & 57.90            \\
DPO             &   72.97         & 86.89           &   79.89         &  93.93              & 85.71               &  88.37        &  78.67         &  84.26         & 83.84            \\
IPO             &   74.92     &   \textbf{88.03}     & 81.36        &  94.32    &  {86.03}             &  \textbf{90.40}       &  80.65     & 85.00      &  85.09          \\
KTO             &  44.46     &  85.04     &   72.52     &  93.35         &  78.61      & 82.74     &  72.27            &  80.80       &  76.22       \\
ORPO            &  63.15     &   86.13       &  79.99     &  93.68        &   74.59       &  89.81         &         78.07    &  85.60        &   81.38       \\
R-DPO           &  25.27       &  84.66      &  76.71      & 59.23             &  80.35            &   83.63     &  72.61       & 80.00        &  70.35          \\
LoPO           &  74.43        & \textbf{88.03}       & {80.91}       &  {94.35}              & {86.03}             &   90.19        & 80.29        &  85.6    & 84.97         \\
RoPO            &  \textbf{74.50}           &  \textbf{88.03}         & \textbf{81.43}           &   \textbf{94.57}                 &     \textbf{86.26}            &   89.98          &   \textbf{81.72}             &    86.93       &   \textbf{85.43}         \\ \bottomrule
\end{tabular}
}
\caption{Accuracy comparison of aligned \textbf{Meta-Llama-3-8B} model with various methods on 8 commonsense reasoning datasets. TS-FT denotes task-specific datasets finetuned model, and the rank(r) of LoPO is 16. The best results are highlighted with \textbf{bold}.}
\label{table2}
\end{table*}

\noindent \textbf{Evaluation.}
We employed three popular and challenging benchmarks to evaluate the open-ended instruction-following task: MT-Bench \cite{zheng2023judgingllmasajudgemtbenchchatbot}, AlpacaEval 2 \cite{alpaca_eval}, and Arena-Hard \cite{arenahardli2024crowdsourceddatahighqualitybenchmarks}. Among them, AlpacaEval 2 consists of 805 questions, and MT-Bench contains 80 questions falling into the following eight common categories: writing, role-play, extraction, reasoning, math, coding, knowledge I (STEM), and knowledge II (humanities/social science). Moreover, Arena-Hard is an extension of MT-Bench, which collects 500 challenging high-quality prompts, achieving a state-of-the-art agreement with human preference rankings.
We present the findings using automatic and LLM-based evaluation methods. For LLM-based evaluation, we employ GPT-4-turbo-2024-04-09  
as the judge model to conduct pairwise comparisons for each preference optimization method and GPT-4. We consider the win rate (WR) against the responses generated by GPT-4 to evaluate the instruction following performance of each method. Since GPT-4 has a tendency to give higher scores to longer responses during evaluation \cite{dubois2024lengthcontrol}, the length should be taken into account simultaneously when comparing performances. Under similar scores, we hold that shorter responses are superior. Therefore, we report the length-weighted win rate (WWR) as the final results by multiplying with the ratio of the content length generated by GPT-4 to that generated by the current method to exclude the influence of length as shown in shown in the following formula. To alleviate positional bias, we assess each candidate in both positions within two separate runs, and the ultimate result is calculated as the average of the two runs.
\begin{equation}
\text{WWR}= \text{WR} \cdot \frac{\text{Len}(y_\text{g4})}{\text{Len}(y)}
\end{equation}
where $y_\text{g4}$ denotes responses generated by GPT-4, and Len denotes length of generation.

For automatic evaluation, we use an open well-tuned reward model\footnote{https://huggingface.co/weqweasdas/RM-Mistral-7B} in RewardBench leaderboard to measure the level of human preference gained, denoted as reward model score (RMS) \cite{xiong2024iterative}. In addition, given that the alignment process might cause damage to the diversity of the generated content, we add corresponding metrics to assess it. Distinct N-grams counts the number of distinct N-grams in the set of outputs \cite{li-etal-2016-diversity}, and we calculated the geometric mean of it with $n = 1,2,3,4$ for measuring diversity.


For the evaluation of commonsense reasoning QA, we regard the accuracy as the metric and compute the mean value across 8 test sets. Overall, we conduct a comprehensive evaluation from four aspects: alignment ability, generation length, generation content diversity, and knowledge forgetting. Among them, the latter three aspects can be regarded as references for assessing alignment overfitting.

\subsection{Experimental Results}
In this section, we present experimental results of different methods, highlighting the superior comprehensive performance of RoPO.
\begin{table*}[]
\centering
\renewcommand\arraystretch{1.0}
\scalebox{1}{
\begin{tabular}{lcccccccccccc}
\toprule
\multirow{3}{*}{\textbf{Method}} & \multicolumn{6}{c}{\textbf{Mistral-Base (7B)}}                                                                              & \multicolumn{6}{c}{\textbf{Llama3-Base (8B)}}                                                                               \\ \cline{2-13} 
                                 & \multicolumn{2}{c}{\textbf{AlpacaEval 2}} & \multicolumn{2}{c}{\textbf{Arena-Hard}} & \multicolumn{2}{c}{\textbf{MT-Bench}} & \multicolumn{2}{c}{\textbf{AlpacaEval 2}} & \multicolumn{2}{c}{\textbf{Arena-Hard}} & \multicolumn{2}{c}{\textbf{MT-Bench}} \\ \cline{2-13} 
                                 & \textbf{WWR}          & \textbf{Len.}      & \textbf{WWR}         & \textbf{Len.}     & \textbf{WWR}        & \textbf{Len.}    & \textbf{WWR}          & \textbf{Len.}      & \textbf{WWR}         & \textbf{Len.}     & \textbf{WWR}        & \textbf{Len.}    \\ \hline
RoPO                             & \textbf{13.03}       & 1038               & \textbf{15.85}      & 1419              & \textbf{22.84}     & 968              & \textbf{11.00}       & 1062               & \textbf{18.63}      & 1509              & \textbf{26.20}     & 1010             \\
single                           & 8.23                 & 847                & 10.54               & 1274              & 11.99              & 863              & 9.38                 & 917                & 11.73               & 1305              & 14.75              & 901              \\
w/o $\underline{m}$              & 9.66                 & 997                & 12.74               & 1398              & 13.28              & 1001             & 8.57                 & 1028               & 14.51               & 1529              & 16.28              & 997              \\
uni-D                            & 11.58                & 1044               & 13.74               & 1432              & {21.96}     & 1007             & 9.84                 & 1041               & 16.45               & 1519              & 22.70              & 974              \\
rot*                             & 10.38                & 1086               & 12.69               & 1453              & 18.10              & 1059             & 9.89                 & 1098               & 15.18               & 1523              & {26.11}     & 1072             \\ \bottomrule
\end{tabular}
}
\caption{Ablation experiment results on three instruction following benchmarks: AlpacaEval 2, Arena-Hard, and MT-Bench.  }
\label{table3}
\end{table*}
\subsubsection{Instruction Following}

Table \ref{table1} reveals that RoPO outperforms all the existing offline preference optimization methods across all benchmarks and settings. On the MT-Bench benchmark, RoPO surpasses the best baseline by 5.1 to 5.3 points, which is even achieved under the premise of lower generation length. In contrast, DPO and KTO achieve marginally higher performances at the cost of significantly increased generation length and reduced diversity. We contend that these methods overfitted in alignment optimization severely. By observing Figure \ref{figure5}, it can be discerned that RoPO achieved the top-2 reward model score on three benchmarks, indicating its decent alignment ability. Additionally, all preference optimization baselines except IPO reduce the diversity of generated content, among which DPO is the most significant, while RoPO improves diversity in comparison with SFT model. Under the experimental setting where Mistral-7B-v0.1 serves as the backbone, the increase in generation length of original DPO, KTO and R-DPO is excessively high. We believe this is attributed to the excessive suppression of dispreferred samples in the original DPO method, resulting in the repetitive patterns in the model's responses, while KTO and R-DPO fail to mitigate it. Notably, RoPO effectively avoids this problem as the generation length of RoPO increases within a reasonable range.

Moreover, compared with adjusting the intensity of the regularization term $\beta$ and optimizing on the loss (such as IPO), RoPO demonstrates significant advantages in suppressing overfitting. Despite the fact that the former two approaches also effectively curb overfitting, the performance degradation on the instruction-following task is intolerable. Nevertheless, RoPO maintains the fundamental expressive capacity and diverse generation capability while acquiring a pretty good alignment ability. Besides, there is no requirement for complicated hyperparameter search efforts for RoPO. In comparison with the normal LoRA-based weight regularization method, RoPO is evidently a more superior weight regularization approach. This is manifested in the better performance and fewer training parameters under a comparable level of generation diversity. We will conduct a detailed analysis of this in the subsequent section. To sum up, we contend that RoPO strikes a well balance between alignment performance and generation diversity.

\subsubsection{Commonsense Reasoning}

To validate the issue of knowledge forgetting caused by the alignment overfitting, we compare the influence of different preference optimization methods on the performance of commonsense reasoning tasks. Table \ref{table2} exhibits that DPO, KTO, and R-DPO resulted in a decline of the model's general ability. By observing some response cases, we discovered that the model appeared to directly answer the content of the options instead of answering the options themselves. It even provided safe response like "Sorry, I don't know." This implies that overfitting during the alignment training lead to the forgetting of task format knowledge and the decline of the model's question understanding ability. In contrast, RoPO still maintain the performance well on Commonsense Reasoning QA, with the accuracy rate on 6 datasets enhanced. We speculate that this is because RoPO retains the knowledge encoded in the angle between neurons well and acquires additional commonsense knowledge during the alignment process.

\subsection{Analysis}
\subsubsection{Ablation Study}
To evaluate the efficacy of RoPO, we carried out ablation studies. In Table \ref{table3}, we present results from ablating each key design of RoPO: (1) modifying Bidirectional Integrated Givens Matrix to unidirectional while keeping the number of parameters unchanged (i.e. {uni-D}); (2) removing the reverse Givens rotation matrix $\tilde{G'}$ (i.e. {single}); (3) removing the magnitude-stretching vector $\underline{m}$  (i.e. w/o $\underline{m}$); (4) rotating the neurons using the BIG Matrix ${h}=\underline{m}\cdot({W^0}\cdot\underline{R'})$ (i.e. rot*). We observe that every design of RoPO is crucial as eliminating each design would result in varying degrees of performance degradation on WWR.
The removal of reverse Givens rotation matrix has the most significant impact on the result, with an average reduction of 6.8 points on WWR. This indicates that although it has been proven that an orthogonal matrix integrated by $d - 1$ Givens rotation matrices is sufficient to fit all rotations \cite{GOFTma2024parameterefficientquasiorthogonalfinetuning}, the actual performance is still limited by the network capacity. Excessive regularization makes the model fail to acquire the instruction-following ability. rot* has an equal number of parameters to RoPO, yet difference lies in the mode of rotation. The experiment results, which indicate an average reduction of 2.5 points on WWR, demonstrate the effectiveness of the regularization that maintains the hyperspherical energy invariant. The outcome achieved by rotating each neuron independently rather than fixing the angle between neurons is worse.

\subsubsection{Training Efficiency}

In addition to its outstanding comprehensive performance, RoPO also has the advantages of low trainable parameters and faster training. Compared with the original DPO, RoPO merely demands 0.0086\% of the trainable parameters. Supposing the weight matrix of the neural network is $\boldsymbol{W}\boldsymbol{\in}\mathbb{R}^{d\times n}$, it is easy to calculate that the training parameter quantity of RoPO is $2(d - 1) + n$. By contrast, the training parameter quantity of the baseline DPO-LoRA is $r \times (d + n)$. In our experimental setup, we apply the trainable matrix to the query vectors and value vectors in the attention mechnism. Assuming that the backbone model employs the common Multi-Head Attention (d = n), then the trainable parameter quantity of RoPO is approximately 3d, the parameter quantity of DPO-LoRA (r = 4) is 8d, and the parameter quantity of DPO-LoRA (r = 16) is 32d. RoPO achieves performance exceeding that of DPO-LoRA with significantly fewer parameters.

\section{Related Work}

With the extensive application of LLMs, how to align with human values has received increasing attention. Once the training details of InstructGPT \cite{ouyang2022traininglanguagemodelsfollow} were disclosed, the advancement of Reinforcement Learning from Human Feedback (RLHF) and its associated technologies has expedited at a rapid pace. 
As the scale of the model continues to expand, conducting full fine-tuning of the pre-trained model on downstream tasks is becoming increasingly challenging. The proposal of Parameter-Efficient Fine-Tuning (PEFT) technology has substantially reduced the training and storage costs, significantly expediting the pace of AI research. More detailed introduction of Preference Optimization and PEFT can be found in the Appendix C.


\section{Conclusion}
In this paper, we proposed RoPO, which is the first attempt to design regularization from the weight-updating perspective, effectively alleviating the overfitting problem in DPO. By merely performing rotational and elongation updates on the neurons, RoPO ensures the hyperspherical energy invariant during the preference optimization process. Extensive experiments demonstrate that RoPO achieves a comprehensively superior performance with an extremely small number of trainable parameters, not only effectively alleviating overfitting but also reducing memory usage during training.

\bibliography{aaai24}

\appendix

\section{Appendix A More Implementation Details}

We discover that hyperparameter tuning is of paramount significance for attaining the optimal performance of preference optimization approaches. Hence, to acquire the supreme performance, we executed a sophisticated hyperparameter search. Below, we exhibit the hyperparameter configurations in the experiment. 

Regarding the SFT training, we train models by utilizing the UltraChat-200k dataset with the subsequent hyperparameters: a learning rate of 1e-6, a batch size of 128, a maximum sequence length of 2048, and a cosine learning rate schedule with 10\% warmup steps for 1 epoch. All the models are trained with an Adam optimizer.

For the preference optimization stage, we train the SFT models using the UltraFeedback dataset with the same hyperparameters as SFT training under the full-parameter settings (DPO, IPO, KTO, ORPO, RDPO).  The learning rate was set as 2e-5 for LoPO and DoPO, 1e-3 for RoPO.  For the method-specific hyperparameters, we searched for the following settings: DPO: $\beta=0.1$, IPO: $\tau=2.0$, KTO: $\lambda_l=\lambda_w=1.0,\beta=0.1$, ORPO: $\lambda=0.1$, RDPO: $\alpha=0.003,\beta=0.1$.

For decoding hyperparameters, we use a sampling decoding strategy to generate responses, with a temperature of 0.95, top-p of 0.7, and top-k of 50.

\section{Appendix B Baselines}
DPO: \citet{rafailov2023directpreferenceoptimizationlanguage} derived it by fitting an implicit reward function through the reparameterization.
IPO: \citet{azar2023generaltheoreticalparadigmunderstand} revised the objective to minimize the disparity between the ratio of log-likelihoods and a given threshold to mitigate the overfitting problem of DPO.
KTO: \citet{KTOethayarajh2024ktomodelalignmentprospect} proposed it to directly maximize the utility of generations instead of maximizing the log-likelihood of preferences.
ORPO: \citet{ORPOhong2024orpomonolithicpreferenceoptimization} integrates a penalty term to preclude the learning of undesirable responses while augmenting the probability of learning preferred ones.
RDPO: \citet{RDPOpark2024disentanglinglengthqualitydirect} attempted to add a length regularization term in the loss function to alleviate the abnormally long generation issue.

\section{Appendix C Related Work}
\subsection{Preference Optimization}
With the extensive application of LLMs, how to align with human values has received increasing attention. Once the training details of InstructGPT \cite{ouyang2022traininglanguagemodelsfollow} were disclosed, the advancement of Reinforcement Learning from Human Feedback (RLHF) and its associated technologies has expedited at a rapid pace. RLHF aims to optimize for the maximum reward through interaction with a reward model trained by the Bradley-Terry (BT) model \cite{BTmodelBradley1952RankAO}, typically with the assistance of reinforcement algorithms such as Proximal Policy Optimization (PPO) \cite{pposchulman2017proximalpolicyoptimizationalgorithms}. Nevertheless, RLHF is confronted with challenges like the instability of reinforcement learning and the sensitivity to hyperparameters. To tackle these issues, recent works have devised some RL-free preference optimization methods. \citet{dong2023raftrewardrankedfinetuning} employs the reward model to rank multiple candidate responses obtained by sampling the policy model and selects the sample with the maximum reward for Supervised Fine-Tuning (SFT). SLiC-HF \cite{zhao2023slichfsequencelikelihoodcalibration} utilizes human preferences as the ranking function and directly realizes alignment on off-policy preference data via the sequence-level contrastive approach. RRHF \cite{yuan2023rrhfrankresponsesalign} scores sampled responses from different sources through a logarithm of conditional probabilities and learns to align these probabilities with human preferences via ranking loss. \citet{rafailov2023directpreferenceoptimizationlanguage} theoretically derived Direct Policy Optimization (DPO) by fitting an implicit reward function through the reparameterization method. DPO is straightforward and effective, significantly lowering the threshold for the alignment of LLMs. Subsequently, numerous works have followed the proposition of DPO: RSO combines the merits of SLic and DPO \cite{RSOliu2024statisticalrejectionsamplingimproves}; IPO \cite{azar2023generaltheoreticalparadigmunderstand} theoretically analyzed how the deficiency of the DPO loss led to the weakening of the strength of the KL-regularization during training and revised the objective to minimize the disparity between the ratio of log-likelihoods and a given threshold; KTO \cite{KTOethayarajh2024ktomodelalignmentprospect}, inspired by prospect theory, endeavors to directly maximize the utility of generations instead of maximizing the log-likelihood of preferences. \citet{CPOxu2024contrastivepreferenceoptimizationpushing} designed CPO, which does not require a reference model and is more parameter-efficient. ORPO \cite{ORPOhong2024orpomonolithicpreferenceoptimization} integrates a penalty term to preclude the learning of undesirable responses while augmenting the probability of learning preferred ones. \citet{RDPOpark2024disentanglinglengthqualitydirect} pointed out that a significant manifestation of overfitting in DPO is the bias of excessively long generated content and attempted to add a length regularization term in the loss function to alleviate this issue. \citet{meng2024simposimplepreferenceoptimization} proposed SimPO, considering the average log-probability of a sequence as the implicit reward; \citet{wang2023reverseklgeneralizingdirect} indicated that the mode-seeking property of reverse KL divergence would decrease the diversity of the generated content and replaced it with superior f-divergences. Different from the above approaches, our work constitutes the first endeavor to incorporate regularization from parameter-updating perspective for enhancing DPO.
\par

\subsection{Parameter-Efficient Fine-Tuning.}
As the scale of the model continues to expand, conducting full fine-tuning of the pre-trained model on downstream tasks is becoming increasingly challenging. The proposal of Parameter-Efficient Fine-Tuning (PEFT) technology has substantially reduced the training and storage costs \cite{gu-etal-2024-light}, significantly expediting the pace of AI research. Currently, there exist three mainstream PEFT approaches \cite{lialin2023scalingscaleupguide}: The first one is adapter tuning, which is accomplished by inserting additional trainable modules into the original model \cite{houlsby2019parameterefficienttransferlearningnlp,liu2022fewshotparameterefficientfinetuningbetter,yang-etal-2023-multi-level}; the second one is prompt tuning, which is achieved by concatenating learnable prefix tokens at the beginning of the input \cite{lester2021powerscaleparameterefficientprompt,liu-etal-2022-p}; the last one is reparameterization tuning, where only the delta of partial model parameters are reparameterized with few trainable parameters \cite{hu2021loralowrankadaptationlarge,aghajanyan2020intrinsicdimensionalityexplainseffectiveness,yang2025dynamicearlyexitreasoning}. The last method can merge the learned delta of model parameters into the original model parameters during inference and does not introduce additional costs at the inference stage. It is commonly held that there are two types of schemes for reparameterization tuning, including LoRA-based methods \cite{gao2024parameterefficientfinetuningdiscretefourier,liu2024doraweightdecomposedlowrankadaptation,yang-etal-2022-take,Wang2022seek} and orthogonal fine-tuning (OFT) \cite{qiu2024controllingtexttoimagediffusionorthogonal,liu2024parameterefficientorthogonalfinetuningbutterfly,GOFTma2024parameterefficientquasiorthogonalfinetuning,yang2024factualitydiversityreconcileddecoding,dai2025sgrpoearlyexitreinforcement}, and our method of adding parameter regularization is relatively similar to the OFT method.

\section{Appendix D Sparse Matrix Multiplication Implementation}
Due to the sparsity of forward BIG Matrix $\tilde{G_1}$, the matrix multiplication between it and the parameter matrix can be quickly implemented in the following equivalent way. The remaining three BIG matrices can be accelerated in a similar way.

\begin{equation*}
\begin{gathered}
\mathbf{R}\cdot\mathbf{x}_m=\begin{bmatrix}x_0\\x_1\\x_2\\x_3\\\vdots\\x_{d-2}\\x_{d-1}\end{bmatrix}\otimes\begin{bmatrix}\cos m\theta_0\\\cos m\theta_0\\\cos m\theta_1\\\cos m\theta_1\\\cos m\theta_1\\\vdots\\\cos m\theta_{d/2-1}\\\cos m\theta_{d/2-1}\end{bmatrix}+\begin{bmatrix}x_1\\x_0\\x_3\\x_2\\\vdots\\x_{d-1}\\x_{d-2}\end{bmatrix}\otimes\begin{bmatrix}-\sin m\theta_0\\\sin m\theta_0\\-\sin m\theta_1\\\sin m\theta_1\\\vdots\\-\sin m\theta_{d/2-1}\\\sin m\theta_{d/2-1}\end{bmatrix}
\end{gathered}
\end{equation*}

\end{document}